\documentclass{article}

\usepackage[preprint]{neurips}
\usepackage[utf8]{inputenc}
\usepackage[T1]{fontenc}
\usepackage{hyperref}
\usepackage{url}
\usepackage{booktabs}
\usepackage{amsfonts}
\usepackage{nicefrac}
\usepackage{microtype}
\usepackage{xcolor}
\usepackage{amsmath}
\usepackage{cleveref}
\usepackage{graphicx}
\usepackage{multirow}
\usepackage{enumitem}

\title{EgoTwin: Dreaming Body and View in First Person}

\author{
    \textbf{Jingqiao Xiu\textsuperscript{1}} \quad
    \textbf{Fangzhou Hong}\textsuperscript{2} \quad
    \textbf{Yicong Li\textsuperscript{1}} \quad
    \textbf{Mengze Li\textsuperscript{3}} \quad \\
    \textbf{Wentao Wang\textsuperscript{4}} \quad
    \textbf{Sirui Han\textsuperscript{3}} \quad
    \textbf{Liang Pan\textsuperscript{4}} \quad
    \textbf{Ziwei Liu\textsuperscript{2}} \quad
    \vspace{4pt} \\
    \textsuperscript{1}{National University of Singapore} \quad
    \textsuperscript{2}{Nanyang Technological University} \quad \\
    \textsuperscript{3}{Hong Kong University of Science and Technology} \quad
    \textsuperscript{4}{Shanghai AI Laboratory} 
    \vspace{8pt} \\
    \url{https://egotwin.pages.dev/}
}

\begin{document}

\maketitle

\begin{figure}[h]
    \centering
    \includegraphics[width=\textwidth]{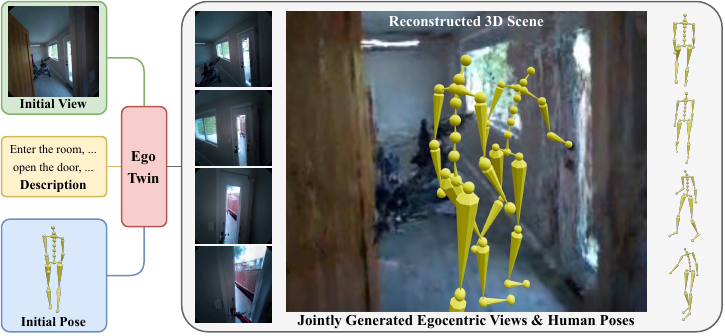}
    \caption{We propose EgoTwin, a diffusion-based framework that jointly generates egocentric video and human motion in a viewpoint consistent and causally coherent manner. Generated videos can be lifted into 3D scenes using camera poses derived from human motion via 3D Gaussian Splatting \cite{kerbl20233d}.}
    \label{fig:teaser}
\end{figure}

\begin{abstract}
While exocentric video synthesis has achieved great progress, egocentric video generation remains largely underexplored, which requires modeling first-person view content along with camera motion patterns induced by the wearer's body movements. To bridge this gap, we introduce a novel task of joint egocentric video and human motion generation, characterized by two key challenges: 1) Viewpoint Alignment: the camera trajectory in the generated video must accurately align with the head trajectory derived from human motion; 2) Causal Interplay: the synthesized human motion must causally align with the observed visual dynamics across adjacent video frames. To address these challenges, we propose EgoTwin, a joint video-motion generation framework built on the diffusion transformer architecture. Specifically, EgoTwin introduces a head-centric motion representation that anchors the human motion to the head joint and incorporates a cybernetics-inspired interaction mechanism that explicitly captures the causal interplay between video and motion within attention operations. For comprehensive evaluation, we curate a large-scale real-world dataset of synchronized text-video-motion triplets and design novel metrics to assess video-motion consistency. Extensive experiments demonstrate the effectiveness of the EgoTwin framework.
\end{abstract}

\section{Introduction}
\vspace{-7pt}
Recent advances in deep generative models have delivered remarkable progress in exocentric (third‑person) video generation \cite{blattmann2023stable,xing2024dynamicrafter,videoworldsimulators2024,yang2025cogvideox}, demonstrating the ability to produce photorealistic and temporally consistent videos from natural language or other conditioning signals. However, egocentric (first‑person) video synthesis remains largely underexplored, despite its increasing importance for wearable computing \cite{fiannaca2014headlock}, augmented reality \cite{ashtari2020creating}, and embodied agents \cite{nair2022r3m}. 
In contrast to exocentric setups, where the camera is static or externally controlled \cite{wang2024motionctrl,he2025cameractrl}, egocentric video captures the perspective of a moving individual, with the footage inherently entangled with the camera wearer's motion. In particular, head movements influence the camera's position and orientation, while full-body actions affect the wearer's body pose and the surrounding scene, collectively shaping the egocentric recording. Therefore, to model body-driven dynamics in egocentric views, we argue that the visual stream must be generated in lockstep with the motion stream that drives it.

In this paper, we introduce a novel task of joint video-motion generation that explicitly models egocentric video together with the motion of the camera wearer. As illustrated in \Cref{fig:teaser}, given a static human pose and an initial scene observation, our goal is to generate synchronized sequences of egocentric video and human motion, guided by the textual description.
This task introduces two fundamental challenges beyond prior works:
\textbf{(1) Viewpoint Alignment.} Throughout the sequence, the camera trajectory captured in egocentric video must precisely align with the head trajectory derived from human motion. This requirement naturally stems from the fact that the camera is rigidly mounted on the wearer's head \cite{engel2023project,applevisionpro2023}, causing head movement and camera motion to be tightly coupled. However, existing exocentric video generation methods typically employ a unidirectional viewpoint-conditioning strategy that synthesizes video based on predefined camera poses \cite{wang2024motionctrl,he2025cameractrl}. Such approaches are unsuitable for our setting, as the camera poses in egocentric video are not externally provided but are inherently determined by the wearer's head motion. As a result, the camera poses must be generated concurrently with the human motion, necessitating a bidirectional interaction to ensure viewpoint alignment.
\textbf{(2) Causal Interplay.} At each time step, the current visual frame provides spatial context that shapes human motion synthesis; conversely, the newly generated motion influences subsequent video frames. Take the ``opening door” scenario in \Cref{fig:teaser} as an example: egocentric observation informs the wearer of the door's location, which guides the wearer's action. In turn, the performed action can alter the body pose (e.g., reaching for the doorknob), the camera pose (e.g., orienting toward the door), and the surrounding scene (e.g., the door gradually opening). These changes must be accurately reflected in subsequent video frames, thereby affecting future motion generation. This recursive dependency forms a closed observation–action loop between video and motion, highlighting the necessity of modeling their causal interplay over time.

To address these challenges, we propose \textbf{EgoTwin}, a joint video-motion generation framework that generates egocentric videos with body-induced camera motion patterns while capturing the causal interplay between visual observations and human actions. Specifically, EgoTwin adopts a diffusion transformer backbone \cite{peebles2023scalable,esser2024scaling}, with three modality-specific branches for text, video, and motion, respectively. To model the joint distribution, EgoTwin employs asynchronous diffusion \cite{bao2023one} in video and motion branches, which allows each modality to evolve on its timestep while maintaining cross-modal interaction.
To facilitate accurate viewpoint alignment, we depart from the commonly used root-centric motion representation \cite{guo2022generating}, which obscures head pose within full-body motion and thus fails to expose the egocentric perspective to the video branch. Instead, we introduce a head-centric motion representation that anchors the human motion to the head joint, allowing for direct alignment between the camera viewpoint of the generated video and the head pose in the synthesized motion.
To faithfully capture the causal interplay, we draw inspiration from the observation-action feedback loop in cybernetic systems \cite{agrawal2016learning,pathak2017curiosity}, where observations shape actions and actions alter future observations. We implement this principle through a structured interaction mechanism: each video token attends to preceding motion tokens, capturing how current observations arise from past actions, while each motion token attends to current and upcoming video tokens, enabling the inference of actions based on perceived scene transitions. This bidirectional design allows motion-driven video synthesis and video-informed motion synthesis to evolve in synchrony.

To foster research in this field, we curate a large-scale dataset of real-world egocentric videos with human pose annotations from Nymeria \cite{ma2024nymeria}. For evaluation, we extend beyond assessing the individual quality of video and motion, and propose video-motion consistency metrics that quantify their cross-modal alignment. Extensive experiments demonstrate the effectiveness of EgoTwin.

In summary, our contributions are fourfold:
\begin{itemize}[leftmargin=10pt]
    \item To the best of our knowledge, we are the first to explore the joint generation of egocentric video and human motion in a viewpoint consistent and causally coherent manner.
    \item We identify the limitations of conventional root-centric motion representations in egocentric contexts and reformulate a head-centric approach that facilitates video-motion alignment.
    \item We design a triple-branch diffusion transformer featuring a video-motion interaction mechanism, supported by an efficient three-stage training paradigm and versatile sampling strategies.
    \item We propose video-motion consistency metrics and build a benchmark for evaluating joint video-motion generation, where our EgoTwin demonstrates strong performance.
\end{itemize}

\section{Related Work}
\textbf{Video Generation.}
Video generation has witnessed significant advancements with the emergence of video diffusion models \cite{ho2020denoising,karras2022elucidating,ho2022video}. A central research focus has been on text-to-video (T2V) generation and image-to-video (I2V) generation, where models synthesize coherent video sequences from textual prompts or static images. Early approaches \cite{blattmann2023stable,blattmann2023align} augment UNet-based text-to-image (T2I) models \cite{rombach2022high} with temporal modeling layers to efficiently transform them to video generation models. Recent works \cite{videoworldsimulators2024,yang2025cogvideox} adopt transformer-based architectures \cite{peebles2023scalable}, achieving improved temporal consistency and generation quality.
To incorporate camera control, representative methods \cite{wang2024motionctrl,he2025cameractrl} inject camera parameters (e.g., extrinsic matrices or Plücker embeddings \cite{sitzmann2021light}) into pretrained video diffusion models \cite{xing2024dynamicrafter,guo2024animatediff}. These approaches rely on known camera trajectories and encode them as input conditions. In contrast, our work considers a fundamentally different setting where the camera trajectory is not available beforehand, yet the generated video must maintain consistency with other synthesized content that is strongly correlated to the underlying camera motion. This key distinction renders existing methods inapplicable, necessitating a framework for controllable video generation that operates without predefined camera guidance.

\textbf{Motion Generation.}
Generating realistic and diverse human motions from text remains a longstanding challenge in computer vision and graphics, offering intuitive control of motion synthesis through natural language. Early works \cite{guo2020action2motion,petrovich2021action,guo2022generating,petrovich2022temos} employ temporal VAEs \cite{kingma2014auto} to capture temporal dependencies and learn probabilistic mappings between language and motion. 
Recent advances have introduced powerful generative modeling techniques to this field, including diffusion models \cite{tevet2023human,zhang2024motiondiffuse,chen2023executing}, autoregressive models \cite{guo2022tm2t,zhang2023generating,jiang2023motiongpt}, and generative masked models \cite{guo2024momask,pinyoanuntapong2024mmm,meng2025rethinking}.
To comply with these frameworks, motion data is represented in different forms. Diffusion-based methods typically operate on continuous vectors, either in the latent space of a VAE or directly from raw motion sequences. Autoregressive models, by contrast, often discretize motion into tokens using vector quantization techniques such as VQ-VAE \cite{van2017neural} or RVQ-VAE \cite{lee2022autoregressive}. Generative masked models are flexible in this regard, accommodating both discrete and continuous representations depending on the loss function and model architecture. 
Furthermore, several researchers \cite{hassan2021stochastic,wang2021synthesizing,huang2023diffusion,zhao2023synthesizing} have investigated human motion generation within 3D scenes represented as RGB point clouds. Others combine the above two tasks by simultaneously incorporating textual and scene information \cite{wang2022humanise,cen2024generating,wang2024move,yi2024generating}. 
Our work differs from this line of research in how scene information is provided: instead of granting full scene access during motion synthesis, we observe the scene only once from the initial human pose and rely on a generative model to hallucinate scene observations as the human moves.

\textbf{Multimodal Generation.}
Recent advances have expanded generative models from unimodal to multimodal generation. Specifically for diffusion models, \cite{ruan2023mm} introduces the first multimodal diffusion framework for synchronized audio-video generation. Other studies \cite{xu2023versatile,bao2023one} design unified models capable of jointly generating text and images. In the domain of human motion, \cite{li2025unimotion} pioneers the simultaneous generation of motion and frame-level language descriptions that explain the generated motions. Despite these developments, the joint modeling of human motion and its corresponding egocentric views remains largely unexplored. To the best of our knowledge, we take the first step in this direction, uncovering the tight coupling between these two modalities.

\begin{figure}
    \centering
    \setlength{\abovecaptionskip}{2pt}
    \setlength{\belowcaptionskip}{-15pt}
    \includegraphics[width=\textwidth]{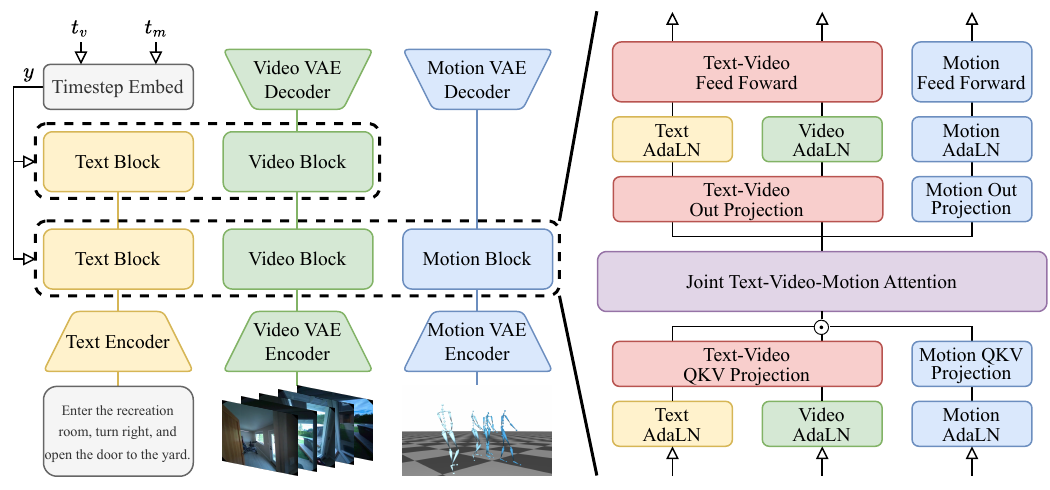}
    \caption{\textbf{Overview.} EgoTwin features a triple-branch architecture \textbf{(left)}, where motion branch spans only the lower half of the layers used by text and video branches. Each branch has its own tokenizer and transformer blocks \textbf{(right)}, with shared weights across branches indicated by matching colors.}
    \label{fig:method}
\end{figure}

\section{Methodology}
\textbf{Problem Definition.}
Given an initial human pose $P^{0} \in \mathbb{R}^{J \times 3}$ in a scene, an egocentric observation $I^{0} \in \mathbb{R}^{H \times W \times 3}$ from that pose, and a textual description of intended human actions in the scene, our goal is to generate two synchronized sequences: (1) a human pose sequence $P^{1:N_m} \in \mathbb{R}^{N_m \times J \times 3}$ and (2) an egocentric view sequence $I^{1:N_v} \in \mathbb{R}^{N_v \times H \times W \times 3}$ spanning the same duration. Here, $J$ is the number of human joints, $H$ and $W$ are the image height and width, $N_m$ and $N_v$ are the number of frames in the pose and view sequences, respectively.
This forms a closed-loop generation paradigm where video and motion mutually and continuously influence each other throughout the sequence.

\textbf{Framework Overview.} An overview of our EgoTwin framework is shown in \Cref{fig:method}. Text, video, and motion inputs are first encoded using a text encoder, a video VAE encoder, and a motion VAE encoder, respectively. These embeddings are then processed through the corresponding branches of a diffusion transformer. Finally, the video and motion outputs are decoded by respective VAE decoders.
\vspace{-15pt}

\subsection{Modality Tokenization}
\label{sec:tokenization}
For the text and video modalities, we adopt T5-XXL \cite{raffel2020exploring} as the text tokenizer and encoder, and a 3D causal VAE \cite{yang2025cogvideox} as the video tokenizer. Specifically, the input text is first tokenized and adjusted to a fixed length $L_t$ via truncation or padding, then encoded into text embeddings $c \in \mathbb{R}^{L_t \times D_t}$. The video frames are temporally and spatially compressed into latent representations $z_v \in \mathbb{R}^{\left(\frac{N_v}{4} + 1\right) \times \frac{H}{8} \times \frac{W}{8} \times C_v}$ with a compression ratio of $4 \times 8 \times 8$ and $C_v$ latent channels, which are subsequently patchified and unfolded into video embeddings $X_v \in \mathbb{R}^{L_v \times D_v}$ of sequence length $L_v$. $D_t$ and $D_v$ denote the embedding dimension of text and video, respectively.

\textbf{Motion Representation.} 
Unlike the uniform representation for text and video, motion representation exhibits a great degree of diversity. Currently, the most widely adopted format in human motion generation is the overparameterized canonical pose representation \cite{guo2022generating}, which has become the default standard for popular datasets, including KIT-ML \cite{plappert2016kit} and HumanML3D \cite{guo2022generating}.
Formally, the human pose at each frame is defined as a tuple of $(\dot{r}^a, \dot{r}^{xz}, r^y, j^p, j^v, j^r, c^f)$, comprising seven groups of features: root angular velocity along Y-axis $\dot{r}^a$, root linear velocities on XZ-plane $\dot{r}^{xz}$, root height $\dot{r}^y$, local joint positions $j^p \in \mathbb{R}^{3(J - 1)}$ and velocities $j^v \in \mathbb{R}^{3(J - 1)}$ in root space, joint rotations $j^r \in \mathbb{R}^{6(J - 1)}$ in local space, and binary foot-ground contacts $c^f \in \mathbb{R}^4$. Motions are retargeted to a default human skeletal template and initially rotated to face the positive Z-axis.

\begin{figure}
    \centering
    \begin{minipage}{0.54\textwidth}
        \centering
        \setlength{\abovecaptionskip}{0pt}
        \includegraphics[height=150pt]{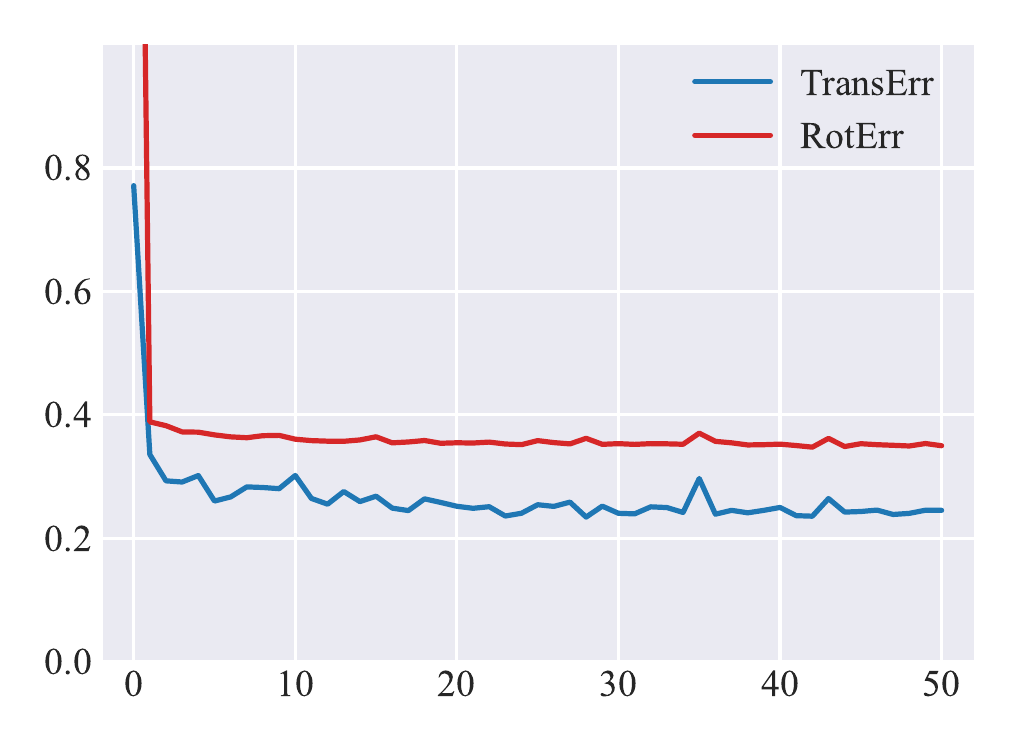}
        \caption{Head pose regression errors over epochs.}
        \label{fig:regress}
    \end{minipage}
    \hfill
    \begin{minipage}{0.45\textwidth}
        \centering
        \setlength{\abovecaptionskip}{0pt}
        \includegraphics[height=150pt]{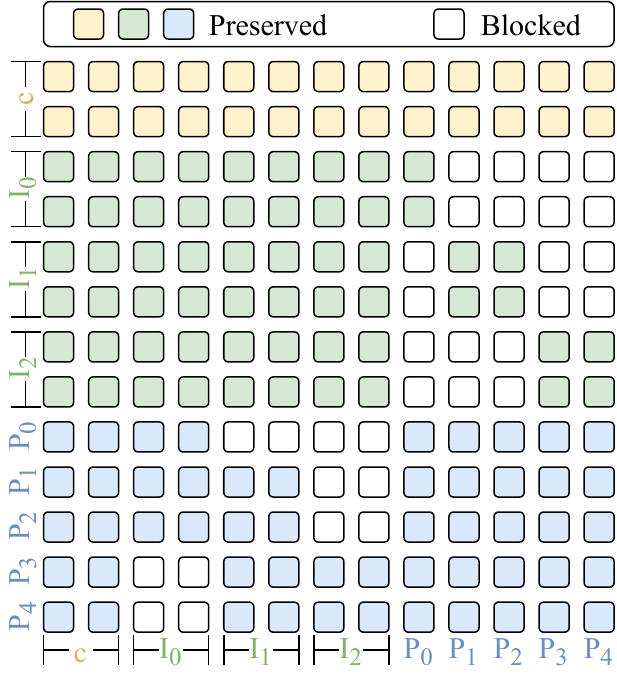}
        \caption{Interaction mechanism.}
        \label{fig:attetion}
    \end{minipage}
    \vspace{-10pt}
\end{figure}

However, the above root-centric representation is not suitable for our task, as the critical information for alignment with egocentric video, such as the pose of the head joint, is deeply buried in an intricate multi-step kinematic calculation. Mathematically, recovering the head joint pose requires integrating root velocities to obtain the root pose, then applying forward kinematics (FK) to propagate transformations through the kinematic chain to the head joint. Intuitively, this computation is too complex to be precisely modeled by neural networks.
To validate our insights, we train a GRU-based regression model that takes motion representation sequences as input, supervised by an MSE loss against ground-truth head pose sequences. As shown in \Cref{fig:regress}, both translation and rotation errors (TransErr and RotErr, see \Cref{sec:metrics} for details) plateau at high levels due to insufficient explicit cues for accurately modeling head pose.

To address this issue, we propose a head-centric motion representation that explicitly exposes egocentric information. Specifically, we define the representation as a tuple $(h^r, \dot{h}^r, h^p, \dot{h}^p, j^p, j^v, j^r)$, where $h^r \in \mathbb{R}^6$ and $\dot{h}^r \in \mathbb{R}^6$ are the absolute and relative rotation of the head joint, $h^p \in \mathbb{R}^3$ and $\dot{h}^p \in \mathbb{R}^3$ are the absolute and relative position of the head joint. The terms $j^p$ and $j^v$ are now expressed in head space, while $j^r$ retains its original meaning. Additionally, we normalize the initial head pose to zero translation and identity rotation, and set all first-order kinematic features to zero in the initial frame. Our representation naturally resonates with egocentric video in at least two novel ways: 1) It offers more accurate access to the head trajectory, which closely correlates with camera movement; 2) It more clearly informs the egocentric video how the body is observed egocentrically.

\textbf{Motion Tokenization.} Inspired by the Causal 3D CNN \cite{yu2024language}, we build the motion VAE using 1D causal convolutions, where all padding is applied at the beginning of the convolutional axis. The encoder and decoder are symmetrically structured, each comprising two stages of $2 \times$ downsampling or upsampling, interleaved with ResNet blocks \cite{he2016deep}. The motion VAE is trained using a combination of reconstruction loss $\mathcal{L}_{rec}$ and Kullback–Leibler (KL) divergence regularization $\mathcal{L}_{\mathrm{KL}}$ weighted by $\lambda_{\mathrm{KL}}$. To ensure that loss contributions are balanced across different groups regardless of their dimensions, we compute the VAE loss $\mathcal{L}_\text{VAE}$ separately for the 3D head ($h^p, \dot{h}^p$), 6D head ($h^r, \dot{h}^r$), 3D joint ($j^p, j^v$), and 6D joint ($j^r$) components. The final loss averages these four items: 
\begin{equation}
    \mathcal{L}_{\mathrm{VAE}} = \frac{1}{4} \sum_{c} \left( \mathcal{L}_{rec}^{(c)} + \lambda_{\mathrm{KL}} \mathcal{L}_{\mathrm{KL}}^{(c)} \right), \text{where} \; c \in \left\{\text{head}_{\text{3D}}, \text{head}_{\text{6D}}, \text{joint}_{\text{3D}}, \text{joint}_{\text{6D}}\right\}.
    \label{eq:vae}
\end{equation}
Using the trained VAE, motion representations are tokenized into latents $Z_m \in \mathbb{R}^{\left(\frac{N_m}{4} + 1\right) \times C_{m}}$ with a $4 \times$ downsampling rate and $C_m$ channels, and subsequently transformed into motion embeddings $X_m \in \mathbb{R}^{L_m \times D_m}$, with $L_m$ as the sequence length and $D_m$ as the embedding dimension.

\subsection{Diffusion Transformer}
\label{sec:transformer}
Our diffusion transformer extends MM-DiT \cite{esser2024scaling}, initially designed for text-to-image generation, to support text, video, and motion modalities.
As illustrated in \Cref{fig:method}, each branch consists of a sequence of MLPs and applies adaptive layer normalization (AdaLN) in conjunction with a gating mechanism \cite{peebles2023scalable} to incorporate timestep information. The text and video branches are initialized from CogVideoX \cite{yang2025cogvideox}, with shared weights except for the AdaLNs. The motion branch corresponds to only the lower half of the layers in other branches, as essential visual cues for video-motion interaction, such as camera pose and scene structure, are primarily captured in the early layers of the video diffusion backbone. In contrast, the higher layers specialize in appearance details, which are less relevant to motion. To further improve efficiency, the motion branch employs reduced channel dimensions, consistent with the lower representational complexity of motion relative to video.
The embedding sequences from different modalities are projected to a common dimensionality $D$ and concatenated for joint attention operations \cite{vaswani2017attention}. This triple-branch architecture allows each modality to work in its own representational space while still attending to and interacting with the others.

\textbf{Interaction Mechanism.}
The original MM-DiT framework includes only text and image modalities, where cross-modal consistency is enforced only at the global level, i.e., matching the entire image with the entire text suffices.
However, our task demands fine-grained temporal synchronization between video and motion: each video frame must be temporally aligned with the corresponding motion frame.
Although we incorporate sinusoidal positional encodings \cite{vaswani2017attention} for both video and motion tokens, along with 3D rotary position embeddings (RoPE) \cite{su2024roformer} for video tokens to provide absolute and relative position information, these mechanisms primarily capture intra-modal structure. 
Consequently, the inter-modal correspondence at each time step remains implicit to the diffusion transformer, which may lead to globally consistent outputs that nevertheless lack frame-wise synchronization.

To address this challenge, we explicitly encode the causal interplay between video and motion by introducing a structured joint attention mask to the diffusion transformer. Given that human motion is typically captured at a higher temporal resolution than egocentric video, we set the number of motion tokens to be twice the number of video tokens (i.e., $N_m = 2N_v$), without loss of generality.
Formally, we follow the notations in Cybernetics \cite{agrawal2016learning,pathak2017curiosity} to rewrite $I^{i}$ as the observation $O^i$, and $(P^{2i + 1}, P^{2i + 2})$ as the (chunked) action $A^i$, where $i \in [0, N_v - 1]$. According to the principles of forward dynamics: $\{O^i, A^i\} \rightarrow O^{i + 1}$ and inverse dynamics: $\{O^i, O^{i + 1}\} \rightarrow A^i$, video tokens corresponding to $O^i$ can attend to motion tokens that correspond to $A^{i-1}$, capturing how $O^i$ arise from $A^{i-1}$, while motion tokens corresponding to $A^i$ can attend to video tokens that correspond to both $O^i$ and $O^{i + 1}$, enabling the inference of $A^i$ based on scene transitions from $O^i$ to $O^{i + 1}$. A special case is given to $P^0$, which is allowed bilateral attention with $I^0$. As demonstrated in \Cref{fig:attetion}, apart from the aforementioned relationship, the remaining attention between video and motion is blocked, while all intra-modal attention, as well as inter-modal attention related to text, are preserved.

\textbf{Asynchronous Diffusion.} We independently sample two timesteps, $t_v$ and $t_m$, between $0$ and $T$ (maximum timestep), and add Gaussian noises $\epsilon_v$ and $\epsilon_m$ associated with these timesteps to the latents $z_v$ and $z_m$, respectively. Each timestep is first encoded via a sinusoidal embedding, and an MLP then processes two concatenated embeddings to produce a unified timestep embedding $y$, which serves as input to the AdaLN layers. Our model consists of a video denoiser $\epsilon_\theta^v(z_v^{t_v}, z_m^{t_m}, c, t_v, t_m)$ and a motion denoiser $\epsilon_\theta^m(z_m^{t_m}, z_v^{t_v}, c, t_m, t_v)$, which are jointly optimized to simultaneously predict the noises added to the video and motion latents using the following objective:
\begin{equation}
    \mathcal{L}_{\mathrm{DiT}} = \mathbb{E}_{\epsilon_v, \epsilon_m, c, t_v, t_m} \left[\left\|\epsilon_v - \epsilon_\theta^v(z_v^{t_v}, z_m^{t_m}, c, t_v, t_m)\right\|_2^2 + \left\|\epsilon_m - \epsilon_\theta^m (z_m^{t_m}, z_v^{t_v}, c, t_m, t_v)\right\|_2^2\right].
    \label{eq:dit}
\end{equation}

\subsection{Training and Sampling}
\label{sec:process}
\textbf{Training Paradigm.} Our training schema comprises three stages: 1) \textit{Motion VAE Training}, as described in \Cref{eq:vae}. 2) \textit{Text-to-Motion Pretraining}. Since the motion branch lacks pretrained weights for initialization, we pretrain it on the text-to-motion task using only text and motion embeddings as input, while keeping the text branch frozen. Following classifier-free guidance (CFG) \cite{ho2022classifier}, we randomly discard the text embeddings with a probability of 10\% to model unconditional motion generation. By omitting the much longer video embeddings at this stage, we can leverage greater parallelism, which accelerates the training process. Critically, freezing the text branch not only preserves the pretrained text-to-video weights but also facilitates the integration of motion embeddings into the pretrained text-video embedding space. 3) \textit{Joint Text-Video-Motion Training}, as formulated in \Cref{eq:dit}. Video embeddings are incorporated in this final stage, and the model learns the joint distribution of video and motion conditioned on text. Again, text embeddings are randomly dropped with a probability of 10\% to model the unconditional video-motion generation.

\textbf{Sampling Strategy.} Benefiting from the joint distribution modeling, our framework supports not only joint video-motion generation conditioned on text (T2VM), but also unimodal generation, including video generation conditioned on text and motion (TM2V), and motion generation conditioned on text and video (TV2M). The CFG for TM2V sampling is defined as follows:
\begin{align}
    \hat{\epsilon}_\theta^v(z_v^t, z_m^0, c, t, 0) = \epsilon_\theta^v(z_v^t, z_m^T, \phi, t, T) + &w_t\left(\epsilon_\theta^v(z_v^t, z_m^T, c, t, T) - \epsilon_\theta^v(z_v^t, z_m^T, \phi, t, T)\right) \notag \\
    + &w_m\left(\epsilon_\theta^v(z_v^t, z_m^0, c, t, 0) - \epsilon_\theta^v(z_v^t, z_m^T, c, t, T)\right).
    \label{eq:cfg}
\end{align}
The CFG formula for TV2M sampling can be derived by exchanging the roles of $v$ and $m$ in \Cref{eq:cfg}. Here, $w_t$, $w_v$, and $w_m$ denote the guidance scales for text, video, and motion conditions, respectively.
For T2VM sampling, taking the motion branch as an example (with the video branch being analogous), its CFG formula is expressed as:
\begin{align}
    \hat{\epsilon}_\theta^m(z_m^t, z_v^t, c, t, t) = \epsilon_\theta^m(z_m^t, z_v^T, \phi, t, T) + &w_t\left(\epsilon_\theta^m(z_m^t, z_v^T, c, t, T) - \epsilon_\theta^m(z_m^t, z_v^T, \phi, t, T)\right) \notag \\
    + &w_v\left(\epsilon_\theta^m(z_m^t, z_v^t, c, t, t) - \epsilon_\theta^m(z_m^t, z_v^T, c, t, T)\right).
\end{align}
After sampling, latents from the video branch are unpatchified to recover their original shape and then decoded by the 3D causal VAE decoder \cite{yang2025cogvideox} to reconstruct the video, while latents from the motion branch are passed through the learned motion VAE decoder to reconstruct the motion.

\vspace{-4pt}
\section{Experiments}
\vspace{-2pt}
\subsection{Evaluation Metrics}
\label{sec:metrics}
\textbf{Video Quality.} We adopt Image Fréchet Inception Distance (\textbf{I-FID}) \cite{heusel2017gans} to evaluate the visual fidelity and realism of individual frames by measuring the distributional distance between the features of generated frames and those of real images. At the video level, we employ Fréchet Video Distance (\textbf{FVD}) \cite{unterthiner2018towards} to quantify temporal coherence and consistency across generated video sequences compared to real ones. Additionally, CLIP Similarity (\textbf{CLIP-SIM}) \cite{wu2021godiva} is utilized to assess the semantic alignment and contextual relevance between generated video clips and textual prompts.

\textbf{Motion Quality.} 
We choose Motion Fréchet Inception Distance (\textbf{M-FID}) \cite{heusel2017gans} to assess the statistical similarity between the high-level features of generated motions and real motions. To evaluate the alignment between text and motion, we train a GRU-based text feature extractor and a GRU-based motion feature extractor, both sharing the same architecture as the evaluator in \cite{guo2022generating}. These models are optimized using a contrastive loss on GloVe \cite{pennington2014glove} text embeddings and our motion representation described in \Cref{sec:tokenization}, ensuring that matched text-motion pairs yield geometrically close feature vectors. Within this learned feature space, the text-to-motion Retrieval Precision (\textbf{R-Prec}) is measured in terms of Top-3 retrieval accuracy. Meanwhile, the Multimodal Distance (\textbf{MM-Dist}) captures the average Euclidean distance between corresponding motion and text features.

\textbf{Video-Motion Consistency.} We propose to evaluate the consistency between generated egocentric videos and human motions from two aspects:
1) \textit{View Consistency}: We first estimate the frame-wise camera poses of the generated egocentric videos using DROID-SLAM \cite{teed2021droid} and extract the head joint poses from the generated human motions. Then, we align both trajectories at the first frame and apply Procrustes Analysis to determine the optimal scale factor that aligns the estimated camera trajectory with the extracted head trajectory. Finally, we compute the Translation Error (\textbf{TransErr}) as the average Euclidean distance between the corresponding camera and head positions, and the Rotation Error (\textbf{RotErr}) as the average angular difference between the corresponding camera and head orientations, using the same formulas as \cite{he2025cameractrl}.
2) \textit{Hand Consistency}: We detect the presence of the left and right hands, equipped with the motion capture device, in the generated egocentric videos. For the generated human motions, we compute the hand visibility from the perspective of a virtual camera mounted on the corresponding head joint with known intrinsics. Based on the presence and visibility analysis, we define the Hand F-Score (\textbf{HandScore}) as the average F-Score of left and right hands, where a \textit{True Positive} means the hand is present in the video and visible from the head in motion, a \textit{False Positive} means the hand is present in the video but invisible from the head in motion, and a \textit{False Negative} means the hand is absent in the video but visible from the head in motion.

\vspace{-2pt}
\subsection{Experimental Setup}
\textbf{Dataset.} To overcome the limitations of using synthetic or small-scale real-world datasets for evaluation, we train and evaluate our model on Nymeria \cite{ma2024nymeria}, a large-scale, real-device dataset that captures diverse people engaged in a wide range of daily activities across various indoor and outdoor locations. The dataset provides paired text-video-motion data, including egocentric videos recorded with Project Aria glasses \cite{engel2023project}, full-body motions captured using the Xsens inertial motion capture system \cite{paulich2018xsens}, and motion narrations written by human annotators. All data are segmented into 5-second clips, yielding approximately 170K samples after filtering, which are split into training, validation, and test sets for the joint training stage. We ensure that both the individuals and environments in the test split remain unseen during joint training.

\textbf{Baseline.}
Since no prior methods are capable of addressing our task, we propose a simple yet effective baseline, VidMLD, that retains the architecture of EgoTwin while removing all dedicated designs introduced in \Cref{sec:tokenization} and \Cref{sec:transformer}. In other words, VidMLD combines the state-of-the-art video diffusion model CogVideoX \cite{yang2025cogvideox} and the latent-space motion diffusion model MLD \cite{chen2023executing}, both of which excel in unimodal generation, and connects them through the multimodal diffusion architecture MM-DiT \cite{esser2024scaling} to enable joint generation. We adopt the same three-stage training recipe described in \Cref{sec:process}, and employ the original classifier-free guidance \cite{ho2022classifier} for sampling.

\textbf{Implementation Details.} In our experiments, videos are undistorted and resized to a resolution of $H = W = 480$, with each segment containing $N_v + 1 = 41$ frames at 8 FPS. The motion data adopts the Xsens skeleton with $J = 23$ joints and consists of $N_m + 1 = 81$ frames per segment at 16 FPS. The video and motion latents have $C_v = 16$ and $C_m = 64$ channels, respectively. The embedding lengths for text, video, and motion are $L_t = 226$, $L_v = 9900$, and $L_m = 21$, with corresponding dimensions $D_t = D_v = D = 3072$, and $D_m = 768$. The hyperparameter $\lambda_{KL}$ in \Cref{eq:vae} is set to 1e-4. CFG scales are set to $w_t = 6$ for text and $w_v = w_m = 4$ for video and motion. The text and video branches have 42 layers, totaling approximately 5B parameters, with most shared across both branches. The motion branch comprises 21 layers, corresponding to the lower halves of the other two branches, and contains roughly 300M parameters.

\subsection{Main Results}
\begin{table}
    \centering
    \setlength{\tabcolsep}{2pt}
    \renewcommand{\arraystretch}{1.2}
    \caption{Quantitative results of joint video and motion generation, evaluated by metrics covering video quality, motion quality, and video-motion consistency.}
    \begin{small}
    \begin{tabular}{lcccccccccc}
    \toprule
    \multirow{2}{*}{\raisebox{-.5\height}{\textbf{Method}}} & \multicolumn{3}{c}{\textbf{Video Quality}} & \multicolumn{3}{c}{\textbf{Motion Quality}} & \multicolumn{3}{c}{\textbf{Video-Motion Consistency}} \\ 
    \cmidrule(r){2-4} \cmidrule(r){5-7} \cmidrule(r){8-10} & I-FID $\downarrow$ & FVD $\downarrow$ & CLIP-SIM $\uparrow$ & M-FID $\downarrow$ & R-Prec $\uparrow$ & MM-Dist $\downarrow$ & TransErr $\downarrow$ & RotErr $\downarrow$ & HandScore $\uparrow$ \\
    \midrule
    \midrule
    VidMLD & 157.86 & 1547.28 & 25.58 & 45.09	& 0.47 & 19.12 & 1.28 & 1.53 & 0.36 \\
    EgoTwin  &  98.17 & 1033.52 & 27.34 & 41.80 & 0.62 & 15.05 & 0.67 & 0.46 & 0.81 \\
    \bottomrule
    \end{tabular}
    \end{small}
    \vspace{-5pt}
    \label{tab:result}
\end{table}
\textbf{Quantitative Results.} As shown in \Cref{tab:result}, EgoTwin significantly outperforms the baseline method across all evaluation metrics, with especially pronounced improvements in video-motion consistency scores. The brute-force joint training of the video and motion generation models leads to poor alignment between the two modalities, resulting in notably lower video-motion consistency performance. In contrast, EgoTwin effectively captures the intrinsic correlation between the two modalities, achieving not only excellent cross-modal consistency but also enhanced single-modal generation quality through the mutually beneficial interaction between video and motion modalities.

\begin{figure}
    \centering
    \includegraphics[width=\textwidth]{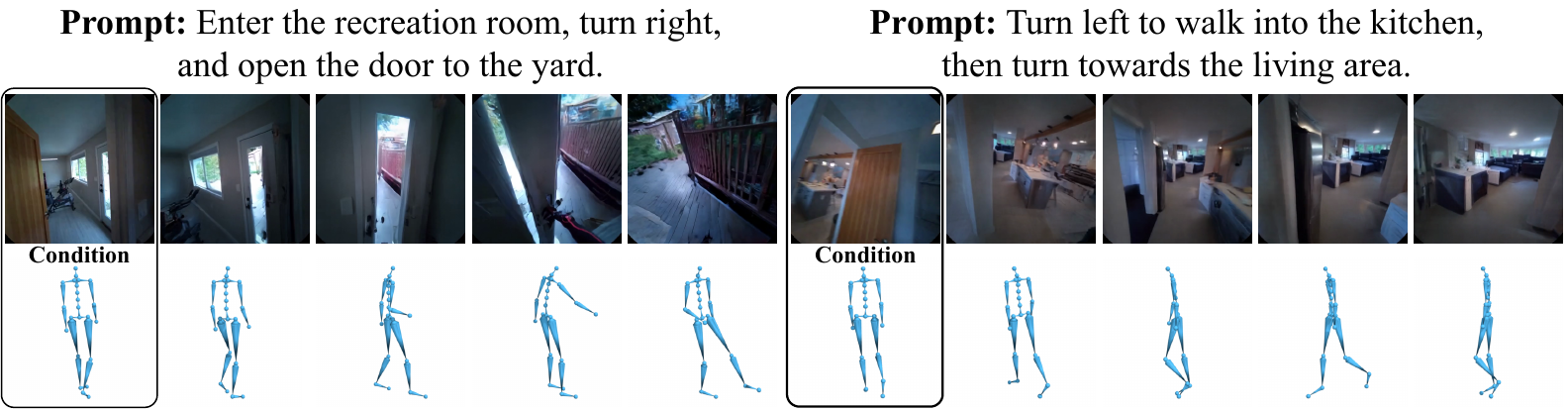}
    \caption{Qualitative results of joint video and motion generation, based on a textual prompt and initial frames of both video and motion.}
    \vspace{-5pt}
    \label{fig:visualization}
\end{figure}

\textbf{Qualitative Results.} We also visualize several examples generated by EgoTwin in \Cref{fig:visualization}. These samples illustrate that both the video and motion streams not only adhere to the textual descriptions for single-modal generation but also evolve in strict cross-modal synchrony, particularly in terms of camera viewpoint and head pose, as well as in scene content and human action. We encourage readers to visit our project page\footnote{\url{https://egotwin.pages.dev/}} for richer generation examples.

\subsection{Ablation Studies}
\begin{table}
    \centering
    \setlength{\tabcolsep}{2pt}
    \renewcommand{\arraystretch}{1.2}
    \caption{Ablation results on three designs: Motion Reformulation (MR), Interaction Mechanism (IM), and Asynchronous Diffusion (AD).}
    \begin{small}
    \begin{tabular}{lcccccccccc}
    \toprule
    \multirow{2}{*}{\raisebox{-.5\height}{\textbf{Variant}}} & \multicolumn{3}{c}{\textbf{Video Quality}} & \multicolumn{3}{c}{\textbf{Motion Quality}} & \multicolumn{3}{c}{\textbf{Video-Motion Consistency}} \\ 
    \cmidrule(r){2-4} \cmidrule(r){5-7} \cmidrule(r){8-10} & I-FID $\downarrow$ & FVD $\downarrow$ & CLIP-SIM $\uparrow$ & M-FID $\downarrow$ & R-Prec $\uparrow$ & MM-Dist $\downarrow$ & TransErr $\downarrow$ & RotErr $\downarrow$ & HandScore $\uparrow$ \\
    \midrule
    \midrule
    w/o MR  & 134.27 & 1356.81 & 26.36 & 43.65 & 0.56 & 17.31 & 0.96 & 1.22 & 0.44 \\
    w/o IM  & 117.54 & 1237.58 & 27.10 & 44.01 & 0.59 & 15.87 & 0.85 & 0.89 & 0.57 \\
    w/o AD  & 109.73 & 1124.19 & 26.91 & 42.58 & 0.53 & 16.48 & 0.74 & 0.62 & 0.73 \\
    EgoTwin &  98.17 & 1033.52 & 27.34 & 41.80 & 0.62 & 15.05 & 0.67 & 0.46 & 0.81 \\
    \bottomrule
    \end{tabular}
    \end{small}
    \vspace{-5pt}
    \label{tab:ablation}
\end{table}
We present the results of extensive ablation studies in \Cref{tab:ablation}, where each row corresponds to a specific ablation setting. All variants exhibit a consistent performance decline across all metrics compared to our full model (listed at the bottom), confirming the effectiveness of each design. First, we replace our Motion Reformulation with the standard representation \cite{guo2022generating} commonly used in human motion generation research (``w/o MR”). The resulting performance drop highlights the importance of our reformulation in exposing egocentric motion cues to the video, which fundamentally facilitates the alignment between egocentric video and human motion. Next, we remove the Interaction Mechanism from the joint attention operations and instead apply full attention without masking (``w/o IM”). The observed degradation underscores its critical role in capturing causal relationships between video and motion, as well as ensuring fine-grained temporal synchronization. Finally, we substitute the Asynchronous Diffusion with a synchronous counterpart for video and motion latents, and accordingly simplify the sampling algorithm to vanilla CFG (``w/o AD”). The performance decline validates its value for modeling comprehensive and diverse dependencies between video and motion modalities, and enabling precise textual control over the joint generation process.

\subsection{Broader Applications}
\begin{figure}
    \centering
    \includegraphics[width=\textwidth]{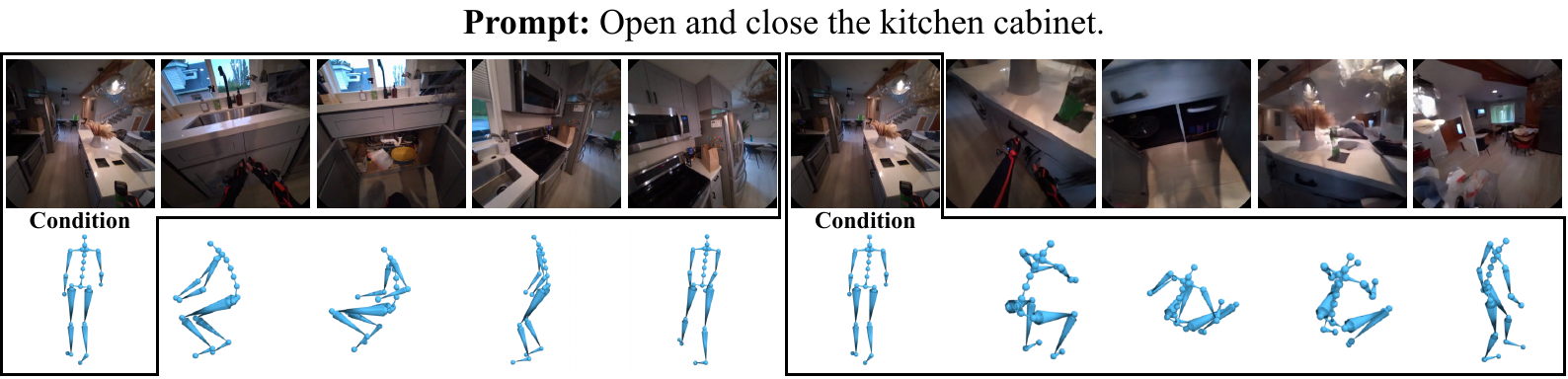}
    \caption{Results of conditional generation. \textbf{Left:} motion generation conditioned on text and video; \textbf{Right:} video generation conditioned on text and motion.}
    \vspace{-10pt}
    \label{fig:application}
\end{figure}

\textbf{Conditional Generation.} Our learned joint distribution enables conditional sampling of one modality given another, using the CFG algorithm described in \Cref{eq:cfg}. As shown in \Cref{fig:application}, we can generate human motion conditioned on text and egocentric video (left), as well as generate egocentric video conditioned on text and human motion (right). Interestingly, textual descriptions are often ambiguous (e.g., they may refer to cabinets on the left or right side of the scene in \Cref{fig:application}), the ability to additionally condition on either motion or video provides greater control over the generation process, which further substantiates the strong coupling between video and motion in our model.

\textbf{Scene Reconstruction.} With jointly generated video and motion, we can effortlessly extract camera poses from human motion and directly integrate both modalities into a 3D Gaussian Splatting \cite{kerbl20233d} pipeline. As illustrated in \Cref{fig:teaser}, we reconstruct the 3D scene from the generated video and seamlessly position the synthesized human into it by aligning head poses with camera trajectories. The realistic spatial interactions exhibited, such as the feet on the ground and the right hand near the door handle, demonstrate strong spatiotemporal coherence between the generated video and motion.

\vspace{-4pt}
\section{Conclusion}
\vspace{-4pt}
We propose EgoTwin, a diffusion-based framework that jointly generates egocentric video and human motion in a viewpoint consistent and causally coherent manner. Our method introduces a head-centric motion representation and a cybernetics-inspired interaction mechanism, supported by an efficient three-stage training paradigm and versatile sampling strategies. To evaluate our approach, we establish a comprehensive benchmark that includes a large-scale dataset of text-video-motion triplets and novel video–motion consistency metrics. Experiments demonstrate that EgoTwin delivers promising results. We hope our work encourages further exploration of joint generative modeling for egocentric video and human motion, and lays a solid foundation for future research in this area.

{
    \small
    \bibliographystyle{abbrv}
    \bibliography{main}
}

\end{document}